\documentclass{article}
\usepackage[final]{neurips_2024}
\usepackage{natbib}
\usepackage{xcolor}
\usepackage{times}
\usepackage{graphicx} 
\usepackage{hyperref}
\usepackage{url}
\usepackage{caption}
\usepackage{subcaption}
\usepackage{amsmath}
\usepackage{amssymb}
\usepackage{comment}
\usepackage{algorithm}
\usepackage{algpseudocode}
\usepackage{booktabs} 
\usepackage{enumitem}

\newtheorem{defn}{Definition}

\newcommand{\cam}[1]{{#1}}
\newcommand{\dejavu}{\emph{d\'{e}j\`{a} vu }}
\newcommand{\Dejavu}{\emph{D\'{e}j\`{a} vu }}
\newcommand{\recall}{\mathsf{recall}}
\newcommand{\precis}{\mathsf{prec}}

\title{\Dejavu Memorization in Vision--Language Models}

\author{%
  Bargav Jayaraman \\
  FAIR, Meta \\
  California, USA \\
  \texttt{bargav@meta.com} \\
  \And
  Chuan Guo \\
  FAIR, Meta \\
  California, USA \\
  \texttt{chuanguo@meta.com} \\
  \And
  Kamalika Chaudhuri \\
  FAIR, Meta \\
  California, USA \\
  \texttt{kamalika@meta.com} \\
}

\begin{document}

\maketitle

\begin{abstract}
Vision-Language Models (VLMs) have emerged as the state-of-the-art representation learning solution, with myriads of downstream applications such as image classification, retrieval and generation. A natural question is whether these models memorize their training data, which also has implications for generalization. We propose a new method for measuring memorization in VLMs, which we call \emph{d\'{e}j\`{a} vu memorization}. For VLMs trained on image-caption pairs, we show that the model indeed retains information about individual objects in the training images beyond what can be inferred from correlations or the image caption. We evaluate \emph{d\'{e}j\`{a} vu} memorization at both sample and population level, and show that it is significant for OpenCLIP trained on as many as 50M image-caption pairs. Finally, we show that text randomization considerably mitigates memorization while only moderately impacting the model's downstream task performance.
\end{abstract}

\section{Introduction}\label{sec:intro}

\begin{figure*}[tb]
    \centering
    \includegraphics[width=\textwidth]{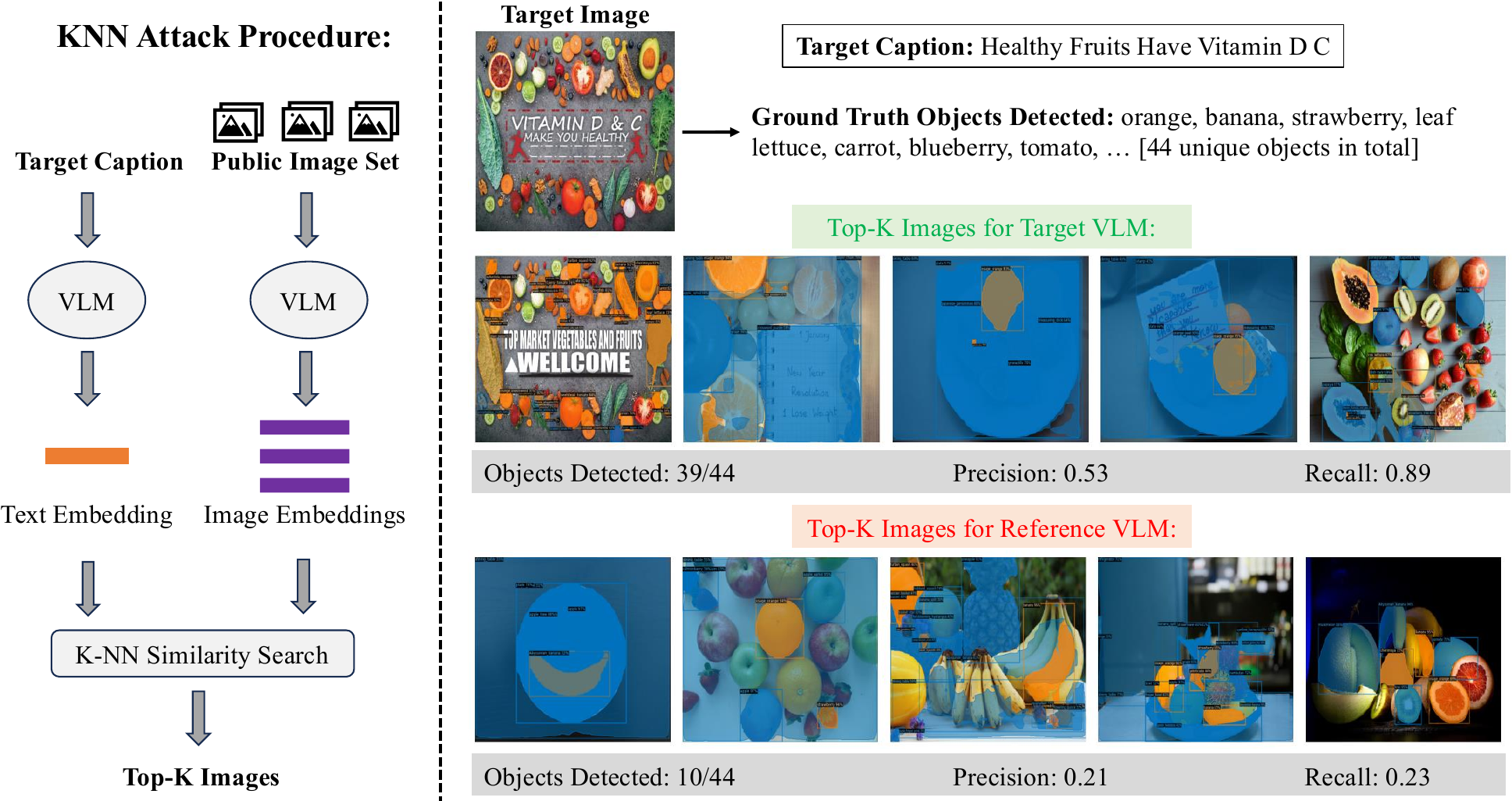}
    \caption{An example where a CLIP~\citep{radford2021learning} model trained on a 40M subset of a Shutterstock data set exhibits \dejavu memorization of objects present in a training image. Public set is a separate collection of 20M images from Shutterstock that has no overlap with the training set. The objects annotated in {\color{orange} orange} are true positives, i.e., the ones present in the target image, and the objects annotated in {\color{blue} blue} are false positives. \cam{Our test recovers significantly more memorized objects for the target VLM (trained on the target image) compared to the reference VLM (not trained on the target image)}. Additional qualitative examples can be found in \autoref{fig:additional_examples_coco} in the appendix.}
    \label{fig:ss_example}
    \vspace{-2ex}
\end{figure*}

Vision-Language Models (VLMs) have emerged as the state-of-the-art solution for learning representations from images and text data, with a number of downstream applications such as image generation~\citep{ramesh2021zero, ramesh2022hierarchical, yu2022scaling}, retrieval~\citep{wang2015deep, cao2016deep, zhang2021privacy, baldrati2022conditioned}, captioning~\citep{mokady2021clipcap}, and classification. At the same time, large foundation models are known to memorize and retain information about their training data~\citep{carlini2019secret, Meehan2023, carlini2023extracting}, and hence, a natural question is whether these Vision-Language Models {\em{memorize}} as well. If so, this raises questions about 
generalizability of these models. We investigate whether Vision-Language Models retain information about their training data beyond the bounds of generalization. 


The main challenge in measuring memorization is designing a measurement technique that can tease apart memorization from spurious correlations. For example, for an image of a black swan on water, a representation learning model may learn to predict {\em{black swan}} given the background {\em{water}} if either: \emph{(i)} it retains extra information about the training image, or, \emph{(ii)} if most of the examples in the training corpus with water also involve black swans. The first kind constitutes as memorization whereas the second kind is spurious correlation. This uncoupling of memorization from spurious correlation is particularly complicated for VLMs. Unlike generative models, VLMs as well as other representation learning models lack decoders that can directly generate images or text; therefore, what the model learns about its training data has to be detected more subtly.

Prior work has looked into this problem for image-only representation models~\citep{Meehan2023} by measuring whether the model can predict the foreground of an image (e.g, black swan) beyond simple correlations based simply on its background (e.g, water). However, such simple solutions do not apply here. VLMs have two separate modalities -- text and image, and the data sets used to train and evaluate them are considerably more complex than the simple foreground-background structure of ImageNet (see \autoref{fig:image_comparison} for an example). A consequence is that the image and text modalities can interact and transfer information in these models in subtly complex ways, making measurement significantly more challenging. 


In this work, we propose a new method for measuring memorization in VLMs (depicted in \autoref{fig:ss_example}). Given a target image caption, we use the VLM to encode the caption and retrieve relevant image samples from a \emph{public set} of images. Our test is based on the key insight that if an image-text pair is memorized by a VLM, then the retrieved images would resemble the training image to a significantly higher amount of detail than what is predictable from either the text caption or simple correlation. Formally, given a text-image pair, we retrieve an image from the model based on an embedding of its text description, and we measure what fraction of ground-truth objects in the original image also co-occur in the retrieved image. Then, to determine whether this happens simply due to correlation, we measure how this compares with the same statistic obtained from a similar VLM which does not have this image-text pair in its training data. Combining these two steps gives us a measurement method that we call VL-D\'{e}j\`{a}-Vu. 

We evaluate our test on CLIP~\citep{radford2021learning} models trained on subsets of Shutterstock and a filtered version of LAION (filtered LAION) with varying number of training samples. We find that even at training data set sizes where CLIP generalizes well, there is a significant degree of model memorization as depicted by our metrics (see \autoref{sec:exp}).
Finally, we explore mitigation measures that reduce information leakage in \autoref{sec:defenses}. We find that text masking significantly mitigates d\'{e}j\`{a} vu memorization at a marginal cost to the model utility. We note that there could be other effective mitigations but were not explored due to the computational limitations. 


\paragraph{Contributions.} Our main contributions are as follows. 

\begin{itemize}[leftmargin=*]

\item We propose VL-D\'{e}j\`{a}-Vu---a new way of measuring memorization in VLMs by measuring what fraction of ground-truth objects in an image can be predicted from its text description for a training image-text pair.

\item Based on this measurement technique, we propose both (a) an individual sample-level test to detect memorization for individual text-image pairs and (b) an aggregate population-level test for a Vision-Language Model. 

\item We use our VL-D\'{e}j\`{a}-Vu test to evaluate memorization in CLIP, and show that memorization does occur for VLMs trained using a number of different training set sizes and regularization parameter values, even for settings where the model generalizes well. 

\item Finally, we explore mitigation measures, and demonstrate that among a number of different ways to train CLIP, random masking of text serves to significantly reduce \dejavu memorization. 

\end{itemize}

\section{Background}

\textbf{Vision-Language models} \citep{radford2021learning, li2022blip, yu2022coca, li2023scaling, xu2023demystifying} are multi-modal models whose core function is to map image-text pairs into a pair of representations that are semantically relevant. These embeddings can then be used for downstream tasks such as image classification, captioning, retrieval and generation. VLMs are composed of a \emph{vision block}, consisting of a convolutional network or a vision transformer, and a \emph{text block}, consisting of a transformer, that produce image and text embeddings respectively from input image-text pairs. Given a trained vision-language model $f$, and an image-text pair $z = \langle z_{img}, z_{txt} \rangle$, we denote the corresponding image and text embeddings as $f(z_{img})$ and $f(z_{txt})$.

We consider VLMs that involve contrastive pre-training; in other words, during training, the model learns to minimize the distance between the image and text embeddings of the matching pairs in the training set $D_{tr}$ while maximizing the distance of the mismatched pairs. The most commonly used contrastive loss is the InfoNCE loss~\citep{oord2018representation} given as follows:
\begin{equation}
    L = -\log \frac{\exp (f(z^i_{img})^\intercal f(z^i_{txt})/\tau)}{\sum_{j} \exp (f(z^i_{img})^\intercal f(z^j_{txt})/\tau)}
\end{equation}
where $\tau$ is the temperature and $z^j, \forall j \neq i$ are negative examples to contrast against. In practice, for each positive example $z^i$, we use all other examples in a training batch as negative examples. 
The most popular VLM of this type is CLIP (Contrastive Language-Image Pre-Training; \citet{radford2021learning}), trained on an undisclosed data set, which achieves competitive out-of-the-box performance across many transfer learning tasks. OpenCLIP~\citep{openclip} has released an open-source implementation of CLIP, and showed that training on a filtered LAION dataset~\citep{schuhmann2021laion} can achieve comparable performance to the original CLIP model. Our work investigates memorization in OpenCLIP. 


\paragraph{Memorization in ML models.} It is well-known that machine learning models can memorize their training data in ways that enable data extraction. This phenomenon has been studied for both language~\citep{carlini2019secret, carlini2021extracting, zanella2020analyzing, jayaraman2022combing} and vision~\citep{somepalli2023diffusion, carlini2023extracting, sablayrolles2018deja, Meehan2023} models. However, all these works only consider the uni-modal setting, and as such the impact of this phenomenon is not clear in the multi-modal settings. Moreover, almost all the prior studies (except \cite{Meehan2023}) focus on generative models -- language or vision -- where measuring memorization is easier because of the presence of a decoder.

Similar to \cite{Meehan2023}, we investigate the setting of representation learning models, where we do not have a decoder and instead only have access to an encoder. Although unlike \cite{Meehan2023}, who considered vision models that capture the relationship between representation of the background of an image (such as water) and the label of its foreground object (such as black swan), we consider settings where the models are trained on more complex data sets that have multiple objects in any given image. Such a simple foreground-background measurement does not directly apply to our setting of Vision Language Models where the two modalities may leak training data in more subtle and complicated ways. Our work builds upon their test, and extends it to VLMs. 
A more detailed background discussion can be found in \autoref{sec:bkg}.
\section{\Dejavu Memorization for Vision-Language Models}

\Dejavu memorization happens when a foundation model retains information about individual training data points beyond what is expected by simple correlation, and allows the recovery of such information during inference time. An example is when an image representation learning model can confidently predict the foreground of a training image based simply on its background~\citep{Meehan2023}, while similar predictions cannot be made for test images. 

In the context of Vision-Language Models, however, measuring \dejavu memorization is not as simple, due to the presence of multiple modalities as well as the complex nature of the training data. Compared to ImageNet, VLMs are trained on vastly more semantically rich data sets with many more objects as well as complicated captions, which may not capture everything in the image -- see \autoref{fig:image_comparison} for an example. This means that the text and image modalities can interact and transfer information in subtly complex ways, making measurement significantly more challenging.

To resolve this challenge, we instead propose to measure whether the ground truth objects in an image can be predicted from the representation of its caption. 
We rely on the intuition that the caption of an image typically does not include all its objects, and hence high confidence recovery of this level of detail implies some form of memorization. If this prediction can be done significantly more accurately when the image is in the training set of a model than when it is in the test, then the image-text pair is being memorized by the said model.

\begin{defn}[\Dejavu Memorization]\label{def:deja_vu}
A vision-language model $f$ suffers from d\'{e}j\`{a} vu memorization if it retains specific information about the individual training images that allows the recovery of objects present in the training images. In other words, for a target image-text pair $z = \langle z_{img}, z_{txt} \rangle$, more unique objects can be recovered from $z_{img}$ given $z_{txt}$ when $z$ is present in $f$'s training set compared to when it is not.
\end{defn}

This is possible due to the model's ability to encode the individual objects in the image embeddings, which is in turn reflected in the corresponding text embeddings when the model minimizes the contrastive loss during training. Next we will discuss how we quantify this phenomenon using two separate models (a target and a reference) as well as a nearest neighbor test.

\subsection{Measurement Methodology}
\label{sec:attack_method}

Since VLMs are meant to capture general correlations between images and their text captions, our goal is to differentiate the recovery of ground-truth objects due to \dejavu memorization from \cam{dataset-level} correlations alone. \cam{As a motivating example, consider the use of CLIP in a cross-modal retrieval task, where images are retrieved from a web-scale database given text. We wish to capture the degree of surprise in the retrieval result when the model memorizes training captions, i.e. how many objects can the model recover beyond dataset-level correlation?} To enable this \cam{evaluation} for a given image-text pair $z = \langle z_{img}, z_{txt} \rangle$, we use two separate VLMs $f_A$ and $f_B$ that are trained on randomly sampled but disjoint data sets $A$ and $B$ respectively. $z$ lies in the training set of {\em{exactly one}} of these models, and hence by comparing the outputs of the two models, we can infer whether $z$ was memorized. 
We do a $k$-nearest neighbor test using a separate public set of images as described in \autoref{alg:knn} and find the subset of images that are closest to $z$ in the representation space. We then decode the objects present in these images. \cam{For this we use an object detector to provide ground-truth annotations for measuring the precision and recall of object recovery. We note that while there will always be some bias when using object detectors, human or automated, this bias should not affect our evaluation when considering the gap between the two models. This is because the object detector is not trained on the same training set as the VLM, hence any incurred bias should be independent of the trained VLMs.}

\begin{algorithm}[tb]
\caption{$k$-Nearest Neighbor Test}\label{alg:knn}
\begin{algorithmic}[1]
\Statex \makebox[.4\textwidth]{} \textbf{Setup Phase}
\State Sample two disjoint data sets $A$ and $B$ consisting of image--text pairs of the form $z = \langle z_{img}, z_{txt} \rangle$ and train models $f_A$ and $f_B$ on the respective data sets.
\State Sample a separate public set of images $P$ that is disjoint from the images in $A$ and $B$.
\State For each image $z_{img}^i \in P$, obtain the corresponding image embeddings from both the models, $f_A(z_{img}^i)$ and $f_B(z_{img}^i)$. 
\Statex \makebox[.4\textwidth]{} \textbf{Testing Phase}
\State Sample a record from $A$ set, $z = \langle z_{img}, z_{txt} \rangle \in A$, and obtain the corresponding text embeddings from both the models, $f_A(z_{txt})$ and $f_B(z_{txt})$.
\State Obtain $k$ public images $N_A \subseteq P$ and $N_B \subseteq P$ closest to $z_{txt}$ in the embedding space for $f_A$ and $f_B$ respectively.
\State Evaluate the gap between the fraction of ground-truth objects detected in the sets $N_A$ and $N_B$.
\end{algorithmic}
\end{algorithm}

\subsection{Metrics}\label{sec:metrics}

\cam{Our memorization metrics are built bottom-up from our notion of deja vu memorization for VLMs. We start from fine-grained \emph{sample-level metrics} to more aggregate \emph{population-level metrics}.} The $k$-nearest neighbor test in \autoref{alg:knn} shows how to obtain predictions of the ground-truth objects given an image; we next use these predictions to develop \cam{the} population-level and sample-level memorization metrics. For our evaluation, we adopt the precision, recall and F-score metrics from the information retrieval literature to quantify the fraction of objects memorized by the models. 

\paragraph{Sample-level metrics.} At the sample level, we evaluate the fraction of ground-truth objects memorized by the target model from a given training image--text pair $z = \langle z_{img}, z_{txt} \rangle$. To do this, we run the nearest neighbor test on both the target and reference models, $f_A$ and $f_B$, to obtain their respective neighbor sets $N_A$ and $N_B$ as per \autoref{alg:knn}. We then calculate the \emph{precision}, \emph{recall} and \emph{F-score} values when identifying the ground truth objects present in $z_{img}$ using $N_A$ and $N_B$ and report the gap between the respective values for both the models. \cam{A positive gap corresponds to the target model memorizing the training sample and the magnitude of the gap indicates the degree of memorization.} The precision, $\precis$, and recall, $\recall$, are given by the following equations ($\forall i \in \{A, B\}$):
\begin{equation}\label{eq:prec_rec}
\begin{aligned}
    \precis(z, f_i) = \frac{\# \text{ unique objects in } N_i \cap z_{img}}{\# \text{ unique objects in } N_i},
    \quad
    \recall(z, f_i) = \frac{\# \text{ unique objects in } N_i \cap z_{img}}{\# \text{ unique objects in } z_{img}}.
\end{aligned}
\end{equation}
F-score is the harmonic mean of precision and recall.

\textbf{Population-level metrics} measure what fraction of the training data is memorized by a model. For proper measurement, we propose three metrics: \emph{population precision gap} (PPG), \emph{population recall gap} (PRG) and \emph{AUC gap} (AUCG). Given the notations defined in \autoref{alg:knn}, the population precision gap is the the fraction of data points from $A$ where $f_A$ has a higher precision in identifying the ground truth objects than $f_B$ minus the fraction of data points where $f_B$ has a higher precision in identifying the ground truth objects than $f_A$. If no memorization occurs, models $f_A$ and $f_B$ should be interchangeable and hence this gap is zero. Formally,
\begin{equation}
\begin{aligned}
    \mathsf{PPG} = \frac{1}{|A|}\Big(|\{z \in A : \precis(z, f_A) > \precis(z, f_B)\}| - |\{z \in A : \precis(z, f_A) < \precis(z, f_B)\}|\Big),
\end{aligned}
\end{equation}
where $|A|$ denotes the size of the set $A$ and $\precis(z, f_A)$ measures the precision of object prediction on $z$ given the model $f_A$ as defined in \autoref{eq:prec_rec}. We define the population recall gap similarly:
\begin{equation}
\begin{aligned}
    \mathsf{PRG} = \frac{1}{|A|}\Big(|\{z \in A : \recall(z, f_A) > \recall(z, f_B)\}| - |\{z \in A : \recall(z, f_A) < \recall(z, f_B)\}|\Big).
\end{aligned}
\end{equation}

\begin{figure*}[tb]
    \centering
    \includegraphics[width=\textwidth]{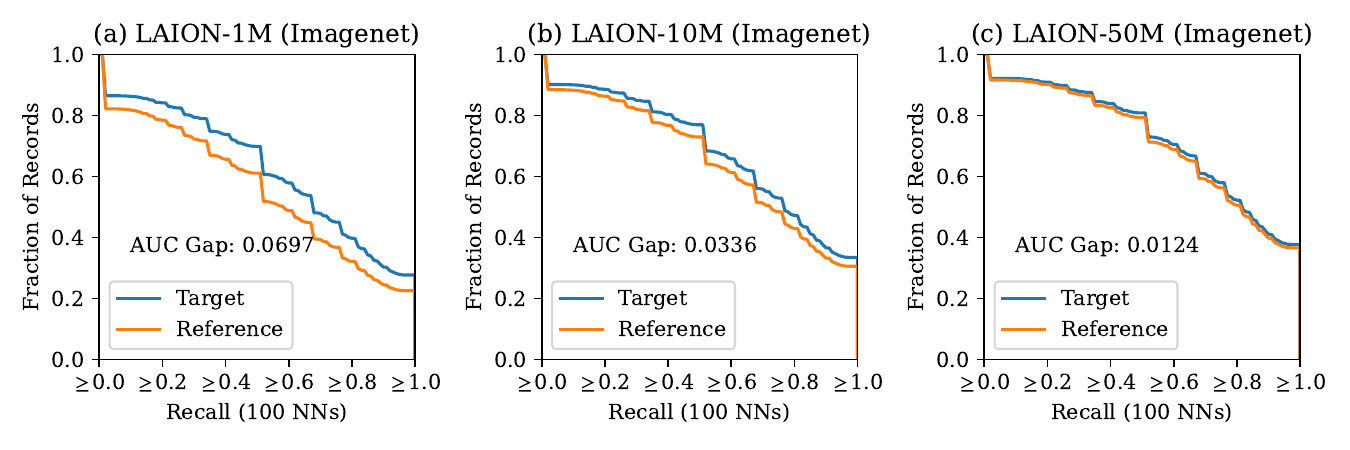}
    \caption{Object recall distribution of target and reference models trained on filtered LAION data set for 200 epochs with different training sizes. ImageNet is used as the public set for kNN test.}
    \label{fig:gap_wrt_datasize}
\end{figure*}

We also visualize the fine-grained cumulative recall distribution of both the models over the training set as shown in \autoref{fig:gap_wrt_datasize}. This gives us a better understanding of what fraction of objects are recovered overall. We then measure the difference between the two distributions (i.e., for $f_A$ and $f_B$) to simplify this information into a single quantity we call AUC gap.

While both the population-level and sample-level metrics rely on the precision and recall functions, they have subtle differences. First, population-level metrics measure the aggregate memorization over the entire training set whereas sample-level metrics measure the memorization in individual training samples. Second, population-level metrics rely on binary tests to differentiate between the target and reference models and as such do not capture the magnitude of the gap between the models as is done by the sample-level metrics. We define both sets of metrics to capture the memorization at different granular levels and to be actionable in a meaningful way, thereby allowing the model developers to fine-tune the models to mitigate the memorization risk.

\section{Evaluating \Dejavu Memorization}\label{sec:exp}

We next apply the metrics designed in Section~\ref{sec:attack_method} to determine if CLIP memorizes training data. Specifically, we seek to answer the following two research questions:

\begin{enumerate}[leftmargin=*]
\item How does d\'{e}j\`{a} vu memorization vary with training set size and number of training epochs?
\item Are all training data points memorized uniformly?
\end{enumerate}

\paragraph{Models and datasets.} We train OpenCLIP from scratch on different datasets, including Shutterstock (a privately licensed data set of 239M image-captions pairs) and $\langle$filtered LAION~\citep{radenovic2023filtering} + COCO~\citep{lin2014microsoft}$\rangle$. We sample up to 50M image-text pairs from the data sets and train OpenCLIP models with ViT-B-32 architecture. For Shutterstock experiments, we consider a separate set of 20M samples from Shutterstock (called SS-20M), with no overlap with the training sets, as public set. For the filtered LAION experiments, we consider two public sets: (a) a separate subset of 50M samples from filtered LAION (called filtered LAION-50M) with no overlap with the training sets, and (b) the entire ImageNet training set~\citep{deng2009imagenet}. More details on the experiment setup and how we obtain data subsets can be found in \autoref{sec:setup}.

\begin{figure*}[t]
    \centering
    \includegraphics[width=\textwidth]{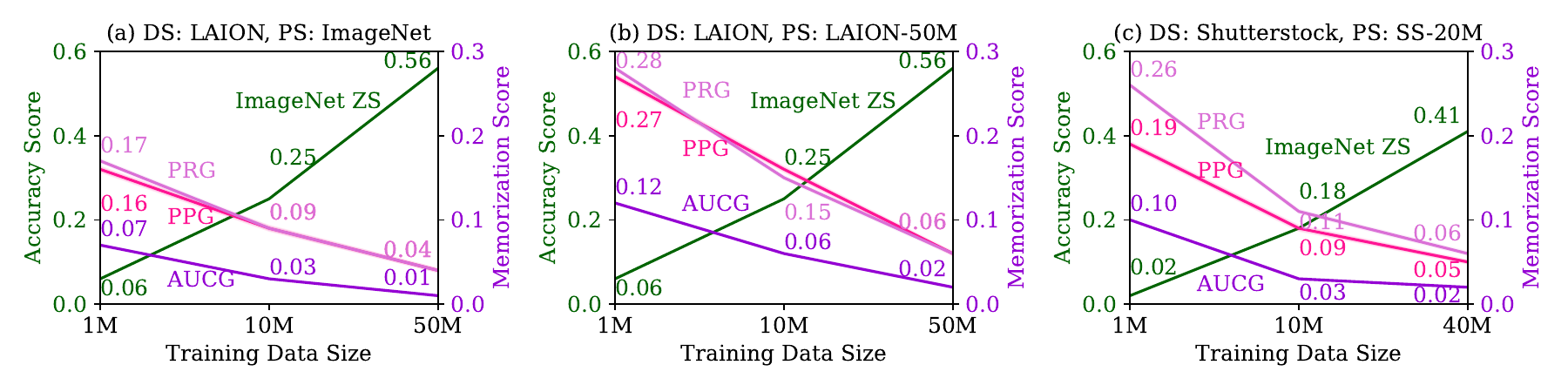}
    \caption{Utility and \dejavu memorization of ViT-B-32 CLIP models with varying training set sizes. Model utility is quantified in terms of ImageNet zero-shot accuracy.   Population-level memorization of models is measured using the metrics defined in \autoref{sec:metrics} over various public sets \emph{(a)}: training set sampled from filtered LAION and ImageNet is used as public set. \emph{(b)}: training set sampled from filtered LAION and a holdout filtered LAION-50M set is used as public set. \emph{(c)}: training set sampled from Shutterstock and a holdout SS-20M set is used as public set. For the memorization metrics, we report the \emph{mean} $\pm$ \emph{std} values (\emph{std} $\le$ 0.003) over 100 repetitions of randomly sampling 10\% of records with replacement.}
    \label{fig:model_acc}
    \vspace{-2ex}
\end{figure*}

\paragraph{Model utility.} As mentioned above (and also discussed in detail in \autoref{sec:setup}), we trained models with different training set sizes consisting of 1M/10M/50M image-text pairs from filtered LAION and 1M/10M/40M image-text pairs from Shutterstock. We use zero-shot performance on ImageNet to evaluate the utility of these models.
\autoref{fig:model_acc} shows the zero-shot accuracy on ImageNet. \cam{Additional utility benchmarks across various ARO (Attribution, Relation, and Order) tasks~\citep{yuksekgonul2023when} can be found in \autoref{fig:vlm_aro} in the appendix.}

\subsection{Measuring Population-Level Memorization}\label{sec:pop}
For quantifying population-level memorization, we measure the gap between the object recall distributions for the target and reference models. 
If there were no memorization, we would observe virtually no gap between the two distributions, i.e. AUCG = 0. \autoref{fig:gap_wrt_datasize} shows the object recall distribution gap between the target and reference models trained on filtered LAION for varying training set sizes when ImageNet is used as the public set. When the training set size is small (e.g. 1M as shown in the left-most figure), there is a higher \dejavu memorization due to the models overfitting on the training set. The gap decreases as the training set size increase from 1M up to 50M, confirming that the models begin to generalize better. Note that the memorization is still significant for models trained on 10M data set. We consider this setting for further experiments as this is a typical training set size for many foundation models in practice \citep{openclip}. For instance, it is common to train CLIP models on the 12M Conceptual Captions data set \citep{cc} or the 15M subset of the YFCC data set \citep{yfcc}. 

Apart from the AUCG \cam{(AUC gap)} metric, we also quantify the gap in terms of the PPG \cam{(population precision gap)} and PRG \cam{(population recall gap)} metrics. \cam{Recall that a positive value for these metrics indicates memorization and the magnitude indicates the degree of memorization.} \autoref{fig:model_acc} shows the PPG, PRG and AUCG metric values for models trained on filtered LAION and Shutterstock with different training set sizes; using ImageNet and filtered LAION-50M public sets for the filtered LAION models and SS-20M public set for the Shutterstock models. Recall that the public sets have no overlap with the model training sets. While the absolute metric values are different for different public sets, the trend remains the same: memorization decreases with increasing training set size as the models begin to generalize better. In \autoref{sec:defenses}, we explore various approaches to reduce this memorization.

\begin{figure*}[tb]
    \centering
    \begin{subfigure}[tb]{\textwidth}
    \centering
        \includegraphics[width=\textwidth]{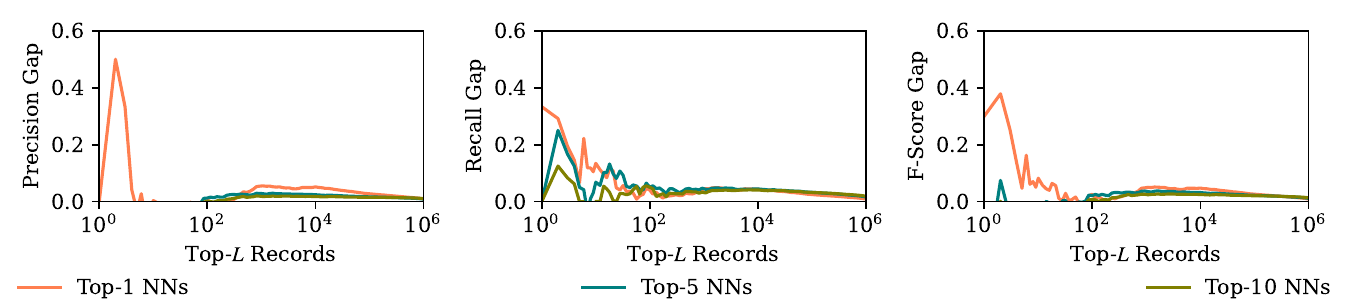}
        \caption{Records are sorted w.r.t. the minimum embedding distance between target caption and public images.}
        \label{fig:lab_inf_top_k_recs_min_dist}    
    \end{subfigure}
    \begin{subfigure}[tb]{\textwidth}
    \centering
        \includegraphics[width=\textwidth]{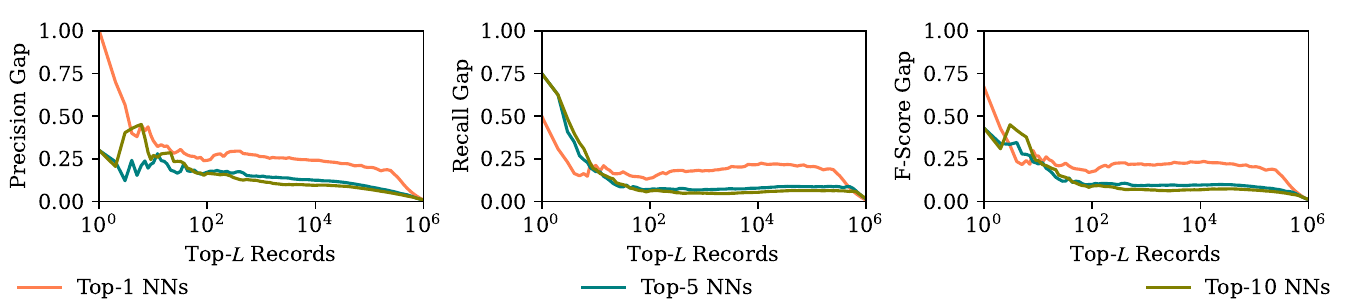}
        \caption{Records are sorted w.r.t. the decreasing number of correct object predictions for target model.}
        \label{fig:lab_inf_top_k_recs_crct_pred}
    \end{subfigure}
    \caption{Sample-level memorization gap between target and reference models when predicting top-10 objects for different top-$L$ records. Models are trained on disjoint 10M subsets of filtered LAION data set for 200 epochs and ImageNet public set is used for the KNN test. The model exhibits very strong \dejavu memorization on a small subset of samples, as indicated by the large precision/recall/F-score gaps when $L$ is small.}
    \label{fig:lab_inf_top_k_recs_imagenet}
    \vspace{-2ex}
\end{figure*}

\subsection{Measuring Sample-Level Memorization}\label{sec:imagenet_sample}
While the population-level metrics like AUCG, PPG and PRG show evidence of memorization, they do not pinpoint which training images are more vulnerable. We sort the training data in decreasing order of memorization to show the subset of most vulnerable records. To do this, we explore several sorting metrics. The most straightforward metric is the distance between the training text embedding and the nearest neighbour public image embeddings obtained using \autoref{alg:knn}. The records for which the public image embeddings are the closest are more easily memorized by the model. Compared to the population-level memorization, where we keep the experiments parameter-free to the best extent, at the sample-level we want to focus on more fine-grained leakage so we choose top-10 object labels to measure the gap instead of predicting all the objects.

\autoref{fig:lab_inf_top_k_recs_min_dist} shows the precision, recall and F-score gaps between the target and reference models for varying top-$k$ records sorted with respect to this distance metric where ImageNet is used as the public set. As shown, the gaps can be greater than 0.3 for top-1 and top-10 records. 
We also tried sorting the records in the decreasing order of the number of objects correctly identified using the target model with the nearest neighbor test. \autoref{fig:lab_inf_top_k_recs_crct_pred} shows the precision, recall and F-score gaps for the records sorted using this metric. We see that the gap can become very significant for the top-1 and top-10 records. Although this metric requires \cam{access to} the ground truth labels, this is still useful to visualize the worst case examples.
Results for sample-level memorization with filtered LAION-50M public set show a similar trend and can be found in \autoref{sec:laion50m_sample}. Sample-level memorization results for Shutterstock experiments can be found in \autoref{sec:additional_results_ss}.

\paragraph{Key Observations.} We show \dejavu memorization at both population and sample levels. At the population-level, where we measure the aggregate memorization of model over the training set, we find that the memorization decreases with an increase in the training set size. This could be attributed to improved model generalization. At the sample-level, we note that the model memorizes disproportionately---a subset of training image-text pairs are memorized more than the others.
\section{Mitigation}\label{sec:defenses}

\begin{figure*}[t]
    \centering
    \includegraphics[width=\linewidth]{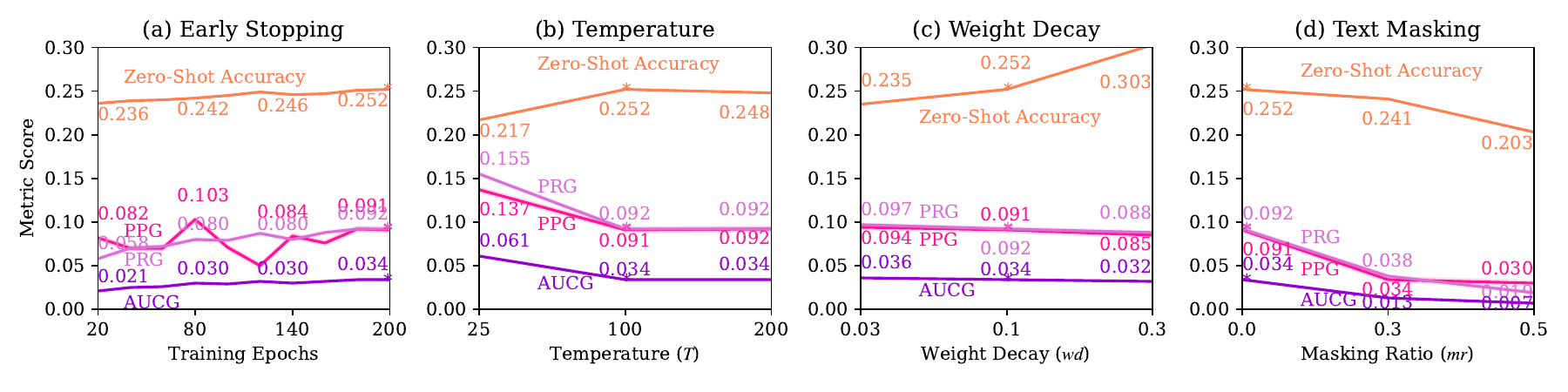}
    \caption{Effect of mitigation on ViT-B-32 OpenCLIP models trained on 10M subset of filtered LAION. Memorization evaluation is done using ImageNet as public set. Default setting is highlighted with asterisk. For the memorization metrics, we report the \emph{mean} $\pm$ \emph{std} values (\emph{std} $\le$ 0.003) over 100 repetitions of randomly sampling 10\% of records with replacement. Among these mitigations, text masking has the best trade-off that reduces memorization without sacrificing utility.}
    \label{fig:defense_impact_imagenet}
    \vspace{-2ex}
\end{figure*}

How can we mitigate \dejavu memorization in VLMs? Since it presumably happens due to the model overfitting on training data, it is likely that regularization techniques may be able to mitigate it. We investigate the impact of four regularization techniques on \dejavu memorization.

\begin{enumerate}[nosep, leftmargin=*]
\item \emph{Early stopping} is a common technique for regularizing neural networks where model training is ended prematurely. It is effective due to the observation that models begin to overfit on the training set when they are trained for more epochs. 
\item \emph{Temperature} is the contrastive loss parameter that controls how close the text and image embeddings can get during the model training. Changing the temperature parameter has a regularization effect for SSL as observed by \citet{Meehan2023}.
\item \emph{Weight decay}, also known as $L_2$ regularization, is a standard ML regularization technique.
\item To reduce overfitting along the text and image modalities in VLMs, we look at additional regularization through {\em{text randomization}}, where we randomly mask a fraction of the text tokens during training. We control the fraction of text tokens masked using a {\em{masking ratio}} parameter. 
\end{enumerate}


In the following we present results when ImageNet is used as the public set for the nearest neighbor test. Results for the filtered LAION-50M public set can be found in \autoref{sec:laion50m_mitigation}. Since Shutterstock memorization trends are similar to those of filtered LAION, we only explore filtered LAION settings for mitigation.

\subsection{Early Stopping}

It is widely believed that deep learning models begin to overfit on the training data as the number of training epochs increases. It is thus a good practice to early stop the training as soon as the model utility on a hold-out test set stagnates or begins to drop. However this is often not the case for SSL models. It is not uncommon to observe that the zero-shot accuracy of SSL models keeps improving as the models are trained for longer~\citep{Meehan2023}. Regardless, we still explore early stopping as a mitigation mechanism. As shown in \autoref{fig:defense_impact_imagenet}, training the CLIP model for more epochs leads to better zero-shot accuracy, but at the same time, \dejavu memorization also increases. This is in line with our hypothesis above. Even when we early stop the model at 20 epochs (10\% of the default parameter value of 200 epochs), the memorization risk is not completely mitigated although the absolute values are lower.

\subsection{Temperature Scaling}

Temperature, or logit scale, controls how close the text and image embeddings can get during training. Smaller values allow for the multi-modal embeddings to get closer, and as a consequence the CLIP contrastive loss drops quickly, whereas larger values regularize the loss but may lead to training instability as noted by \citet{radford2021learning}. The default value in OpenCLIP implementation is set to 100. We vary this value between 25, 100 and 200. As shown in \autoref{fig:defense_impact_imagenet}, decreasing the temperature ($T$) from 100 to 25 decreases the model's zero-shot classification accuracy on ImageNet from 25.2\% to 21.7\% and also increases the memorization as indicated by the increase in the PPG, PRG and AUCG metrics. This is due to the decrease in the distance between the text and image embeddings for the training data which could potentially lead to model overfitting. Increasing the temperature to 200 moderately impacts the model's zero-shot classification accuracy and the memorization leakage remains more or less the same.

\subsection{Weight Decay}
Weight decay directly controls the model overfitting, with larger values corresponding to stronger regularization. The default value is set to 0.1 and we vary it between 0.03, 0.1 and 0.3. As expected, decreasing the weight decay $wd$ from 0.1 to 0.03 decreases the model's zero-shot classification accuracy and also worsens the leakage due to memorization as shown in \autoref{fig:defense_impact_imagenet}. Interestingly, increasing the weight decay to 0.3 significantly improves the model's zero-shot accuracy. We believe that the default value of 0.1 is not optimal for the 10M training set size as it was set based on the model training for larger data sizes (possibly on the entire filtered LAION data set). With 0.3 weight decay, we observe a consistent decrease in the population memorization leakage, as shown by the PPG, PRG and AUCG values for $wd=0.3$ in \autoref{fig:defense_impact_imagenet}, but the values are still significantly high. We also explored setting weight decay to 0.01 and 1.0, but they either adversely impacted the model utility or severely increased memorization. Thus while tuning $wd$ does not completely mitigate memorization, we can get a reasonable trade-off in the neighbourhood of $wd=0.3$.

\subsection{Text Randomization}
During model training, the CLIP models increase the cosine similarity between the matching image-caption pairs while simultaneously decreasing the cosine similarity between mismatched pairs to reduce the contrastive loss. While it is common to augment the training images to reduce overfitting, the text captions are not randomized. This could lead to the model overfitting on the text captions when minimizing the contrastive loss. To avoid this, we propose text randomization as a defense. For COCO subset of the training set, we randomly choose one out of the five captions for each image per epoch during training. For filtered LAION subset, we randomly mask a fraction of caption tokens since only a single caption is available per image in the filtered LAION data set. We vary the masking ratio between 0 (no masking), 0.3 and 0.5 (randomly mask half of the tokens).

We find this defense to work the best in mitigating d\'{e}j\`{a} vu memorization but at the cost of \cam{ImageNet zero-shot accuracy}. As shown in \autoref{fig:defense_impact_imagenet}, using a masking ratio of 0.3 reduces the ImageNet zero-shot accuracy from 25.2\% (in the default case when $mr = 0.0$) to 24.1\%, but at the same time this significantly reduces memorization. The PPG metric reduces from 9.1\% to 3.4\%, and the PRG metric reduces from 9.2\% to 3.8\%. Moreover, the recall CDF gap (AUCG) also reduces from 0.034 to 0.013. Further increasing the masking ratio to 0.5 mitigates the risk even more. PPG reduces to 3.0\%, PRG reduces to 1.9\%, and AUCG reduces to only 0.007. \cam{However, we note that text masking has a positive impact on ARO benchmark utility as shown in \autoref{fig:vlm_aro}. This is because masking avoids overfitting on specific text tokens making the models less likely to behave like bag-of-words. Thus text masking achieves the best utility trade-offs.} We would expect a significant drop in the model utility if we further increase $mr$ since the captions would have considerably less information. 


\paragraph{Key Observations.} We study the impact of tuning four regularization parameters: number of training epochs, temperature, weight decay and masking ratio. We find that early stopping reduces memorization but at the cost of model utility. Increasing the temperature increases the model zero-shot accuracy and decreases memorization up to a certain threshold, beyond which the model utility begins to decrease. Surprisingly, we find that the default value of 100 already gives the optimal results. Similar to temperature, increasing the weight decay increases the model utility and decreases the memorization up to a certain threshold. We find 0.3 weight decay to achieve the best results for a model trained over 10M data. We observe a sharp decrease in model utility beyond this value. Text masking seems to be most effective in mitigating memorization. Increasing the masking ratio decreases memorization and also decreases the model utility. Masking ratio of 0.3 achieves a good trade-off by significantly reducing memorization while only moderately impacting the model utility.

\section{Discussion}\label{sec:discussion}
Prior works have mainly shown memorization in the uni-modal setting: either for the language models~\citep{carlini2019secret} or for vision models~\citep{Meehan2023}. We have demonstrated that even in the complex multi-modal setting, ML models suffer from memorization. Moreover, while prior works have only evaluated memorization for small training data sizes (typically on the scale of 1 million or less), we show memorization on a wide scale, from 1 million to 50 million training set size. Our experiments show that while the population-level memorization metrics decrease with increase in the training set size, there remain strongly memorized examples as exemplified by the sample-level memorization where the model disproportionately memorizes a subset of records.

Careful tuning of right hyper-parameters can, however, mitigate this memorization risk. We propose a suite of metrics to quantify \dejavu memorization in hope of guiding ML practitioners to train models in a safe way. These metrics not only quantify the risk in a meaningful and interpretable manner, but are also sensitive to the tuning of the mitigation parameters, thereby aiding the practitioners in choosing the right model hyper-parameter values that achieve a good utility-risk trade-off. 

\cam{Below we discuss some detailed discussions and limitations of our work.}

\paragraph{\cam{Not applicable to out-of-box models.}}
Since our tests require access to two models, \emph{target} and \emph{reference}, along with the underlying training set, we note that this can not be directly applied to measure memorization in out-of-the-box pre-trained models as there is no reference model for such cases. We leave this case as a future work.

\paragraph{\cam{Distinguishing memorization from learning.}}
\cam{A model can memorize and generalize (or learn) at the same time. This can happen at a sub-population level, where the model memorizes rare concepts and generalizes to common concepts, or even at a sample level, where memorization is required for learning rare concepts as theorized in \cite{feldman2020does}. \Dejavu memorization is meant to go beyond this, and instead examine when a model that is trained on an image with a generic caption (i.e., they do not describe the image in high detail), memorizes many small details about the associated image (i.e., what objects are present in the image) when given the caption. In other words, we define \dejavu memorization as what can be inferred about the training image from its caption beyond simple correlations, which can happen through both learning and memorization in the traditional sense.}

\paragraph{\cam{Extending beyond objects.}}
\cam{While our approach is also applicable to annotations that go beyond objects, this is not in the scope of this work. Even in this setting, the prior state-of-art approach~\citep{Meehan2023} only considers a single object label per image (ImageNet) and none of the prior works consider a. multimodal setting, b. large training size sizes, and c. multiple objects per image.}

\paragraph{\cam{Relation to Overfitting.}} \cam{\Dejavu memorization measures overfitting at a more granular level--- instead of a binary decision, it measures to what \emph{degree} the model overfits a training sample.}


\section*{\cam{Acknowledgements}}
\cam{We thank Diane Bouchacourt for helpful feedback. We would also like to thank Amro Abbas for helping in obtaining the de-duplicated version of filtered LAION data set and Evgenia Rusak for helping with the OpenCLIP implementation.}

\bibliography{ref}
\bibliographystyle{plainnat}

\clearpage
\appendix
\section{License of the assets}
\label{sec:licenses}

\subsection{License for the code}
The licensing information for OpenCLIP~\citep{openclip} can be found at \url{https://github.com/mlfoundations/open_clip/blob/main/LICENSE}.
We use the code from \citet{Meehan2023} for memorization quantification, the licensing information can be found at \url{https://github.com/facebookresearch/DejaVu?tab=License-1-ov-file#readme}. For object annotations, we use Detic~\citep{zhou2022detecting}, the licensing information can be found at \url{https://github.com/facebookresearch/Detic/blob/main/LICENSE}.

\subsection{License for the data sets}
We use ImageNet~\citep{yang2021imagenetfaces} for which the license can be found at \url{https://www.image-net.org/download.php}. We use a filtered version of LAION~\citep{radenovic2023filtering} (which we call filtered LAION) for which licensing information can be found at \url{https://github.com/facebookresearch/diht/blob/main/ LICENSE}. The licensing information for the MS COCO data set~\citep{lin2014microsoft} that we use can be found at \url{https://cocodataset.org/#termsofuse}. We also use Shutterstock data set which is a private licensed data set consisting of 239M image-caption pairs.
\section{Background and Related Work}\label{sec:bkg}



Foundation models, such as large language models, have been long known to memorize their training data in ways that enable easy extraction. For example, a line of work~\citep{carlini2019secret, carlini2021extracting, zanella2020analyzing, jayaraman2022combing} has shown that large language models exactly memorize sequences of text tokens from the training data, and these text tokens can be extracted. \citet{somepalli2023diffusion, carlini2023extracting} showed that diffusion models can generate images that are semantically and stylistically similar to training images or even their near-exact copies under certain circumstances. However, almost all prior studies that demonstrate this kind of memorization focus on generative models -- language or vision -- where measuring memorization is easier because of the presence of a decoder. In contrast, our work is concerned with representation learning models, where we simply have an encoder. 

\citet{sablayrolles2018deja} study d\'{e}j\`{a} vu~\footnote{We note that many prior works have used the term ``d\'{e}j\`{a} vu'' in different contexts. \citet{dhamija2000deja} use this to refer to the ability of humans to recognize images, and they use it as a proxy for password-based authentication. \citet{sablayrolles2018deja} denote d\'{e}j\`{a} vu to essentially mean membership inference, where they test if a model \emph{remembers} if an image was used in training. \citet{Meehan2023} refer to d\'{e}j\`{a} vu as the ability of inferring foreground objects from vision models given a background patch of pixels. We use this term to refer to a vision--language model's ability to recall the individual objects in the training images.} memorization in neural networks and show that it is possible to infer whether an image or a subset of images was used in model training. Our work is also closely related to~\citet{Meehan2023}, which measures d\'{e}j\`{a} vu memorization in image representation models. They show that given the representation of the background of an image, (such as water), the label of its foreground object (such as black swan) can be predicted reliably. Moreover, this prediction is significantly more accurate for images in the training set of a model, thus showing that the models memorize their training data beyond the bounds of spurious correlation. However, such a simple foreground-background measurement does not directly apply to the more complex, multi-modal Vision Language Models where the two modalities may leak training data in more subtle and complicated ways. Our work builds upon their test, and extends it to VLMs. 

Finally, there has been a body of work on empirical measurement of privacy, and broadly speaking, there are three main kinds of attacks. 
In membership inference~\citep{shokri2017membership}, the goal is to determine if a specific data point was used to train a model. In attribute inference~\citep{yeom2018privacy}, the goal is to infer unknown features or attributes of a data point based on a model trained on this or similar points. Finally, training data reconstruction attacks~\citep{fredrikson2015model} aim to recover one or more training data points given a model and some auxiliary information. Our work falls within the purview of attribute inference. However, unlike most attribute inference attacks which were shown to be forms of statistical imputation~\citep{jayaraman2022are}, our tests directly measure how much more effective attribute inference can be when a data point is in the training set of a model.

\section{Detailed Experiment Setup}\label{sec:setup}

For our experiments we use OpenCLIP \citep{openclip} to train the models. For filtered LAION experiments, we train models over subsets of filtered LAION \citep{radenovic2023filtering} and MS-COCO \citep{lin2014microsoft} data sets. For Shutterstock experiments, we train models over various subsets of Shutterstock data set, a privately licensed dataset of 239M image-captions pairs.

\paragraph{Obtaining Data Splits.} As discussed in \autoref{alg:knn}, our test requires disjoint training sets $A$ and $B$ to train the models $f_A$ and $f_B$ respectively, and additionally we require a public set $P$, that has no overlap with $A$ and $B$, for our nearest neighbor search. Moreover, for our tests to be meaningful we need to remove duplicate image--caption pairs otherwise the kNN test becomes trivial if the same sample is also present in the public set and as a result we would overestimate memorization. Conversely, if the same sample is present in both $A$ and $B$ sets, then it is harded to distinguish the outputs of two models and we would underestimate memorization. This type of duplication is common in internet-scraped data sets such as filtered LAION and Shutterstock. We perform semantic deduplication over filtered LAION data set using the procedure of \cite{abbas2023semdedup} to obtain 220M deduplicated image--caption pairs. The Shutterstock data set has different type of duplicates--- multiple images are present with same verbatim captions. So we do a simpler yet effective deduplication by considering only one unique image per caption. This reduces the overall data set size to around 103M image-caption pairs.

\paragraph{- For filtered LAION experiments:} To obtain the two non-overlapping training sets for filtered LAION experiments, we sample 40K image--text pairs from COCO data set and 1M/10M/50M image--text pairs from filtered LAION data set to form $A$ set. We do the same from the remaining pool of data to obtain the $B$ set. Since the COCO part of the $A$ and $B$ sets is insignificant compared to the filtered LAION portion of the sets, we only count the filtered LAION portion size for simplicity when we say we sample 1M/10M/50M training sets. To obtain the filtered LAION-50M public set, we sample 50M pairs from the remaining pool of deduplicated filtered LAION which has 120M pairs (after removing the largest $A$ and $B$ sets from the original 220M data set). We include most of the results on this public set in \autoref{sec:additional_results_laion_50m}. Since this data set may contain human faces, we perform face-blurring on all the sets. We also take the 1.28M images from ImageNet data set~\citep{deng2009imagenet} and perform face-blurring to form our ImageNet public set.

\paragraph{- For Shutterstock experiments:} We take the caption-level deduplicated data set consisting of 103M image--caption pairs and randomly split it into 40M + 40M + 20M sets. The first two 40M sets are used to obtain the 1M/10M/40M $A$ and $B$ sets respectively. The last 20M set is used as the public set. A small portion of the remaining 3M data is used as a hold-out set for hyper-parameter tuning during model training.

\paragraph{Model Hyper-Parameter Settings.} We use the ViT-B-32 CLIP model architecture consisting of around 151M trainable parameters and train the models for 200 epochs using Adam~\citep{kingma2017adam} optimizer with cosine learning rate scheduler and a learning rate of 0.0005. For filtered LAION experiments, we use 256 Nvidia Quadro GP100 GPUs with 16GB VRAM to train the models in parallel with an effective batch size of 16\,384. We set the weight decay to 0.1 and use 1000 warmup steps for the learning rate scheduler. For Shutterstock experiments, we use 32 Nvidia A100 GPUs with 80GB VRAM to train the models in parallel with an effective batch size of 32\,768. We set the weight decay to 0.2 and warmup to 2000 steps. All the model training runs use 512GB RAM and the training time scales with the data size: training on 10M data size takes around 2 days and training on 50M data size takes around 10 days to complete. All other hyper-parameters are set to the default value as used in OpenCLIP; we do an ablation study on the impact of temperature and weight decay in \autoref{sec:defenses}.

\begin{figure*}
    \centering
    \begin{subfigure}{0.39\textwidth}
        \subfloat[][ImageNet Sample.]{\includegraphics[width=0.66\textwidth]{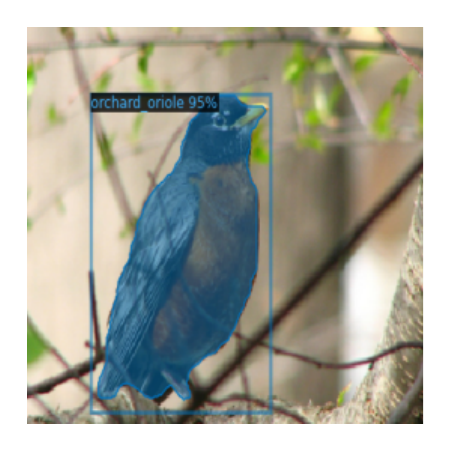}}
        \begin{minipage}{\linewidth}\vskip-80pt \small
        \textbf{Label:}\\ Orchard Oriole.
        \end{minipage}
    \end{subfigure}
    \begin{subfigure}{0.59\textwidth}
        \subfloat[][COCO Sample.]{\includegraphics[width=0.44\textwidth]{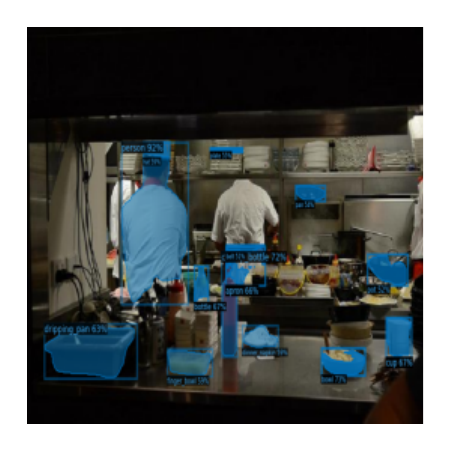}}
        \begin{minipage}{\linewidth}\vskip-90pt \small
        \textbf{Caption:}\\ Several kitchen workers making dishes \\in commercial kitchen.\\
        \textbf{Labels:}\\ Catsup Bottle, Pot, Hat, Plate, Dinner \\Napkin, Finger Bowl, Soda Can, Bottle, \\Dripping Pan, Cup, Work Shirt, Bowl, \\Apron, Person, Belt, Pan.
        \end{minipage}
    \end{subfigure}
    \caption{Comparing images from ImageNet and COCO data sets. The ImageNet images only have single label per image but COCO images have complex scenes with multiple object labels. Additionally, COCO images have accompanying text captions. Label annotations with bounding boxes are highlighted in {\color{blue} blue} for both the images.}
    \label{fig:image_comparison}
\end{figure*}

\paragraph{Obtaining Object Annotations.} For quantitative evaluation of our nearest neighbor tests, we require detailed object annotations for the $A$, $B$ and $P$ sets. Both Shutterstock and filtered LAION data sets only have image captions and no object annotations. ImageNet originally has only one object annotation per image, as shown in \autoref{fig:image_comparison}. Hence, we use an open-source annotation tool, called Detic~\citep{zhou2022detecting}, to obtain multiple fine-grained object annotations per image for all our data sets. This tool can annotate all the 21K ImageNet objects. Detic uses a default threshold of 0.5 to identify object bounding boxes (i.e., any bounding box that has more than 0.5 confidence is considered for annotation). For Shutterstock we use 0.3 threshold as the 0.5 threshold results in nearly 17\% images with no annotations. For all other data sets, we use the default value of 0.5. Even though COCO has multiple object annotations, its class label space is small (i.e., only 80 unique classes). Hence we use Detic on COCO to extend its annotations and to make the label annotations consistent across all the data sets we use. \autoref{fig:ss_example} shows the sample images with multiple object annotations obtained using Detic.

\paragraph{Limitations in Experimental Evaluation.}
We find that the object annotation tool, Detic~\citep{zhou2022detecting}, is not always accurate. For instance, the tool often classifies a `polar bear' as `jaguarundi'. However, our experiments rely on the relative gap in the object detection between the target and reference models and as such are robust to these inaccuracies as long as the annotations are consistent across the images. For instance, if the `polar bear' is classified as `jaguarundi' across all the public set images, the gap between the `polar bear' detection accuracy of target and reference models, based on our nearest neighbor test, will remain consistent. While the absolute numbers in our quantitative tests may vary based on the object annotation tool used, our experimental observations would not change.

\section{Additional Results with filtered LAION-50M}\label{sec:additional_results_laion_50m}

\begin{figure}[tb]
    \centering
    \begin{subfigure}[tb]{\textwidth}
    \centering
    \includegraphics[width=\linewidth]{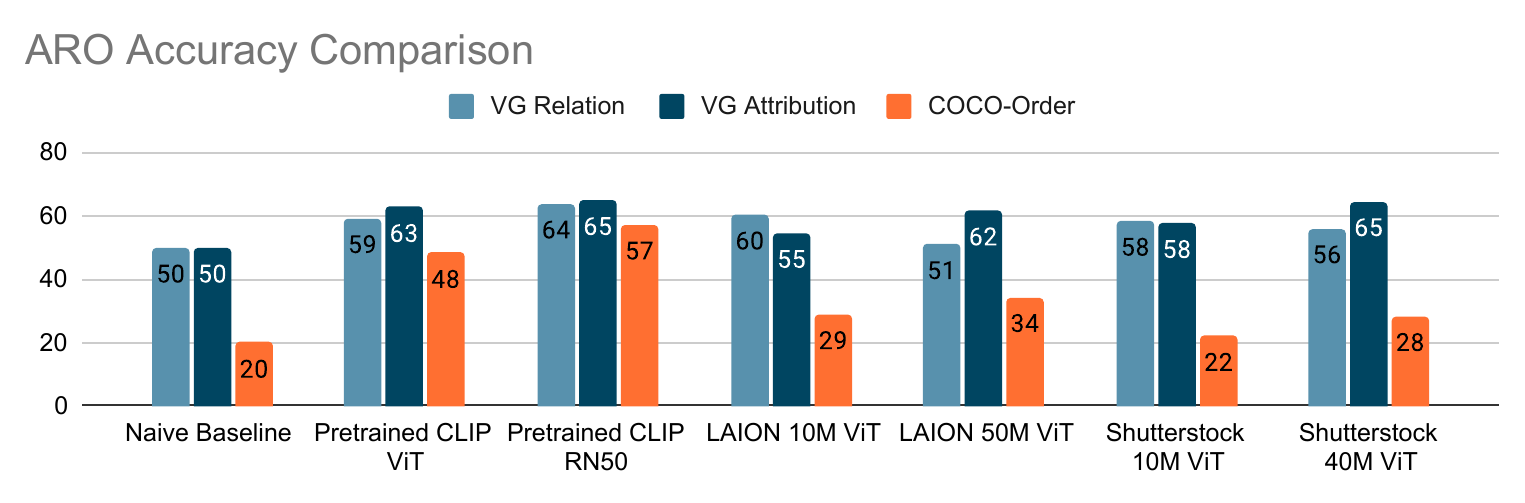}
    \end{subfigure}
    \begin{subfigure}[tb]{0.48\textwidth}
    \centering
    \includegraphics[width=\linewidth]{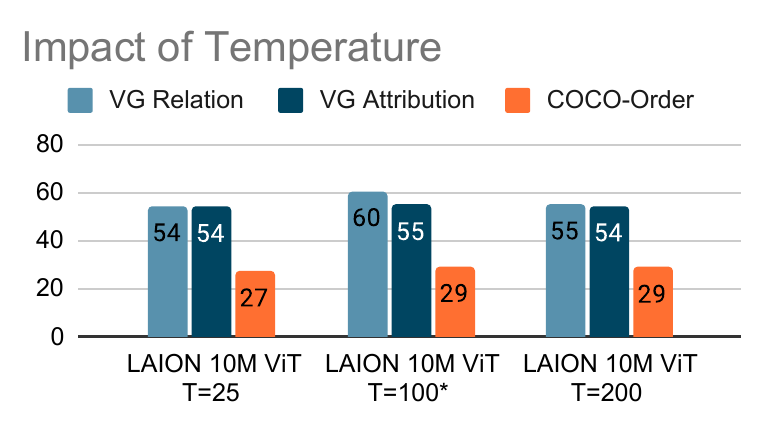}
    \end{subfigure}
    \begin{subfigure}[tb]{0.48\textwidth}
    \centering
    \includegraphics[width=\linewidth]{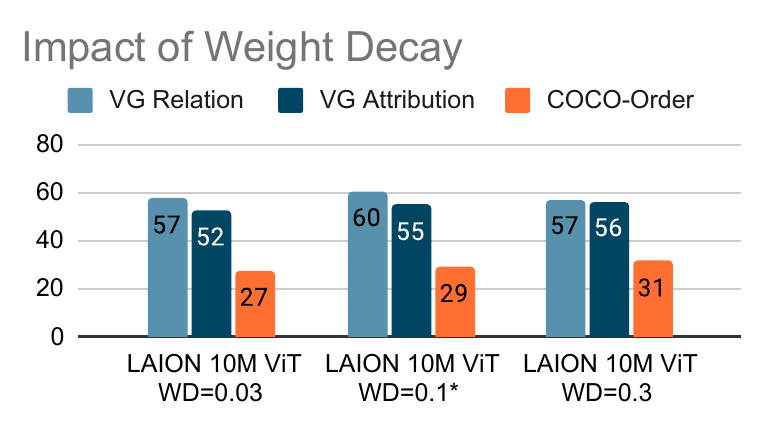}
    \end{subfigure}
    \begin{subfigure}[tb]{0.48\textwidth}
    \centering
    \includegraphics[width=\linewidth]{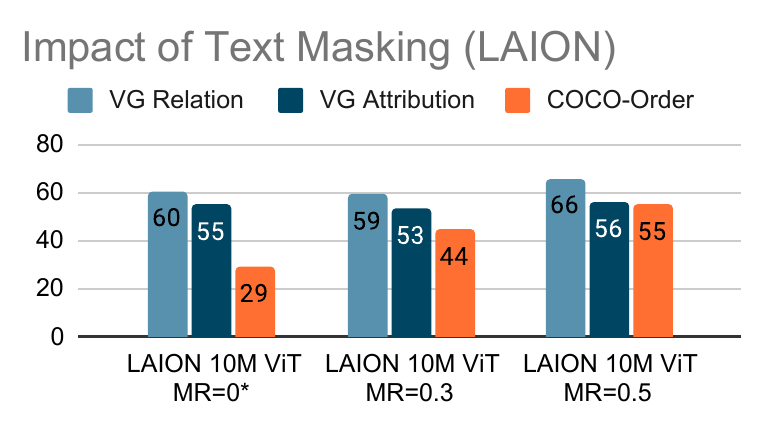}
    \end{subfigure}
    \begin{subfigure}[tb]{0.48\textwidth}
    \centering
    \includegraphics[width=\linewidth]{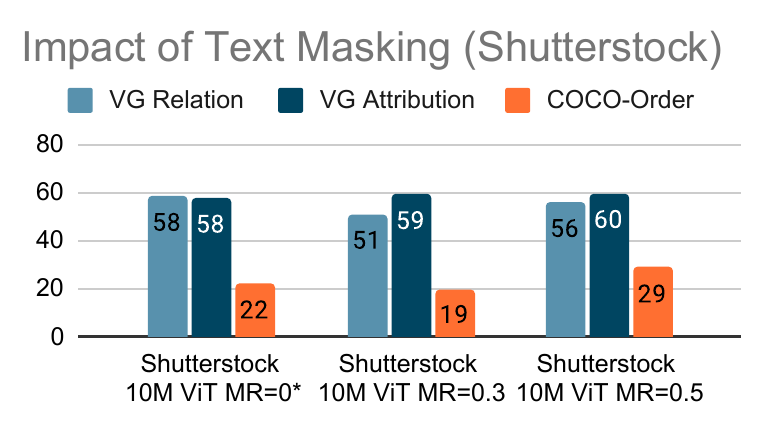}
    \end{subfigure}
    \caption{ARO benchmark accuracy comparison for various models. Top figure compares the accuracy of various baseline models on three compositional reasoning tasks: Visual Genome Attribution, Visual Genome Relation and COCO Order. The pretrained CLIP models are trained on 400M private dataset, whereas our LAION and Shutterstock models are trained on smaller subsets of the respective datasets. Our models are comparable to the pretrained CLIP models in VG Relation and Attribution Tasks. The middle figures show the impact of temperature (left) and weight decay (right) on the ARO accuracy for our models trained on 10M subset of filtered LAION dataset. As shown, the default parameter values (shown by asterisk) achieve the best values for most cases. The bottom figures show the impact of text masking on ARO accuracy for our models trained on 10M subsets of filtered LAION (left) and Shutterstock (right) datasets. Our text masking does not deteriorate the model utility, and in fact further boosts ARO accuracy for COCO ordering task. This is because text masking avoids overfitting on specific text tokens. Thus, unlike the unmitigated CLIP models, the mitigated models are less likely to behave like bag-of-words.}
    \label{fig:vlm_aro}
\end{figure}

In \autoref{sec:exp}, we discussed the memorization results considering the ImageNet as the public set for our nearest neighbor test. Here we discuss the results with a much larger filtered LAION-50M data set as the public set. While the overall trend remains the same as with the ImageNet, with a richer public set, we are able to achieve a larger memorization gap for our models.

\subsection{Sample-Level Memorization}\label{sec:laion50m_sample}
\begin{figure*}[tb]
    \centering
    \begin{subfigure}[tb]{\textwidth}
    \centering
        \includegraphics[width=\textwidth]{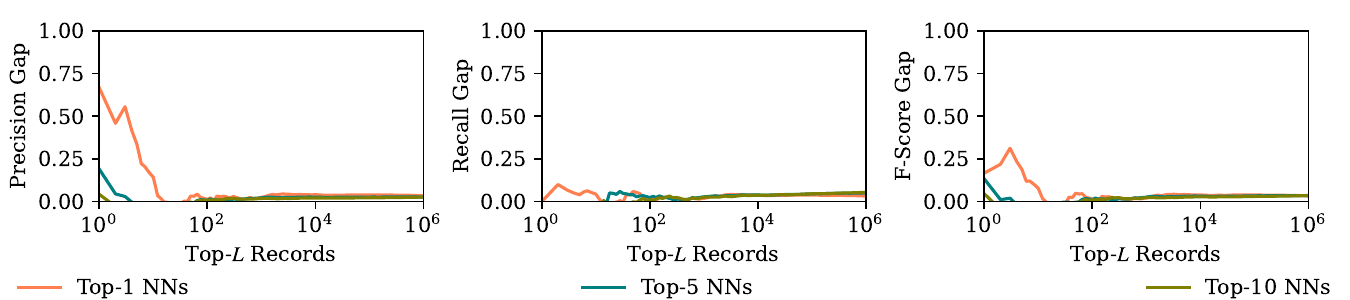}
        \caption{Records are sorted w.r.t. the minimum embedding distance between target caption and public images.}
        \label{fig:lab_inf_top_k_recs_laion_min_dist}    
    \end{subfigure}
    \begin{subfigure}[tb]{\textwidth}
    \centering
        \includegraphics[width=\textwidth]{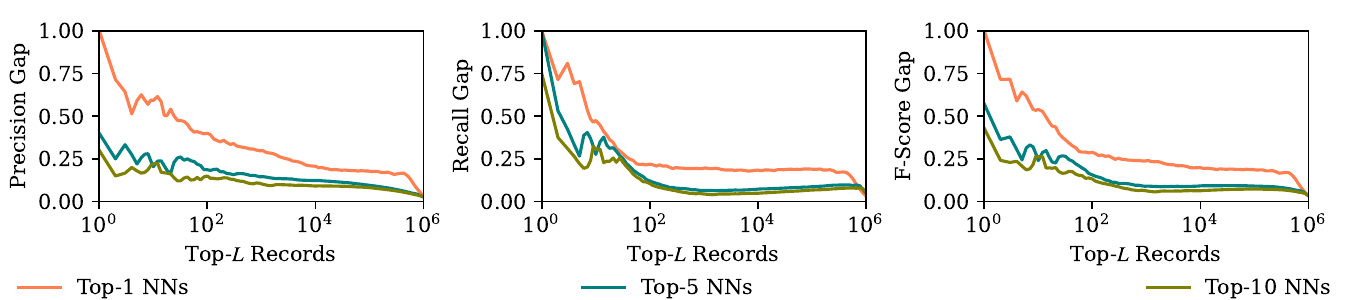}
        \caption{Records are sorted w.r.t. the decreasing number of correct object predictions for target model.}
        \label{fig:lab_inf_top_k_recs_laion_crct_pred}
    \end{subfigure}
    \caption{Sample-level memorization gap between target and reference models when predicting top-10 objects for different top-$L$ records. Models are trained on disjoint 10M subsets of filtered LAION data set for 200 epochs and filtered LAION-50M public set is used for the KNN test.}
    \label{fig:lab_inf_top_k_recs_laion}
\end{figure*}

Similar to the sample-level evaluation for ImageNet public set in \autoref{sec:imagenet_sample}, we evaluate the gap in precision, recall and F-scores of top-$k$ records sorted with respect to the minimum embedding distance when considering filtered LAION-50M as the public set for the nearest neighbor test. \autoref{fig:lab_inf_top_k_recs_laion_min_dist} shows the memorization gap of the top-$k$ records. We note a greater precision gap with top-1 nearest neighbor when compared to the case where ImageNet was used as a public set (see \autoref{fig:lab_inf_top_k_recs_min_dist}). However, the recall gap is lower with this public set. These variations could be due to the nature of the public set--- many filtered LAION images have few or no annotations. This does not mean that the sample-level memorization risk is lower. As shown in \autoref{fig:lab_inf_top_k_recs_laion_crct_pred}, the memorization gap is much higher for this public set when we sort the records in the decreasing order of the number of correct predictions made by the target model using the nearest neighbor test. This corroborates our population-level memorization results in \autoref{fig:model_acc} where we find a higher memorization gap with filtered LAION-50M public set.

\subsection{Mitigation}\label{sec:laion50m_mitigation}
\begin{figure*}[tb]
    \centering
    \includegraphics[width=\linewidth]{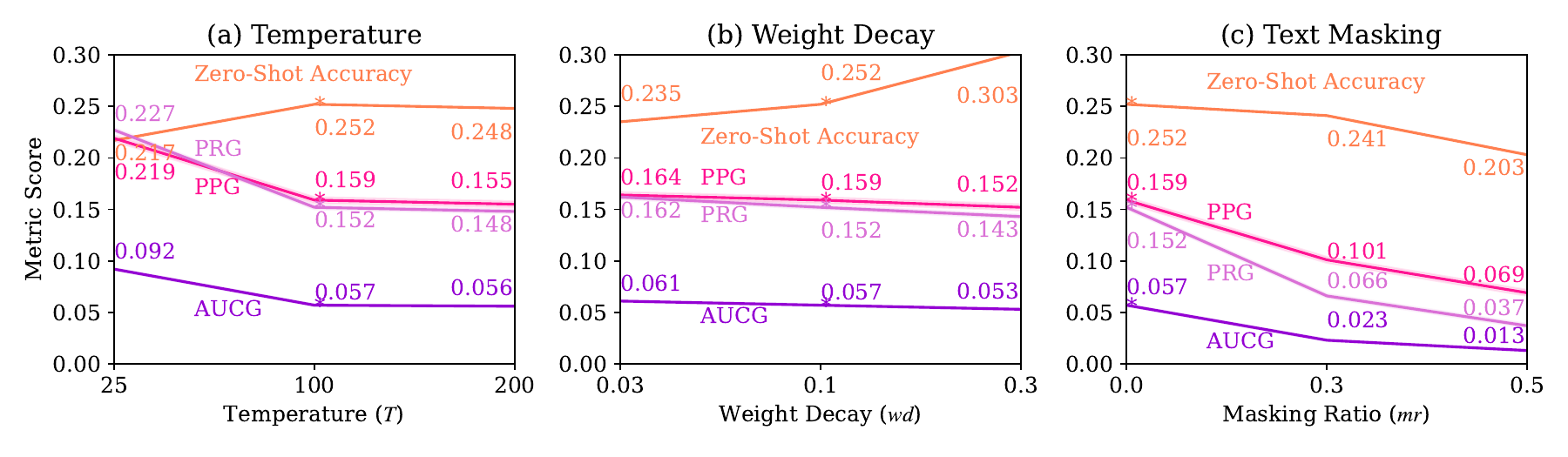}
    \caption{Effect of parameter tuning on ViT-B-32 CLIP models trained on 10M subset of filtered LAION for 200 epochs. Memorization evaluation is done using filtered LAION-50M as public set. Default setting is highlighted with asterisk. For the memorization metrics, we report the \emph{mean} $\pm$ \emph{std} values (\emph{std} $\le$ 0.003) over 100 repetitions of randomly sampling 10\% of records with replacement.}
    \label{fig:defense_impact_laion}
\end{figure*}

We observe similar trends for mitigation with different regularization parameters as with the ImageNet case. \autoref{fig:defense_impact_laion} shows the impact of different parameters on the memorization. Since the filtered LAION-50M public set is much larger than the ImageNet public set, the overall memorization values are higher due to the public set nearest neighbors being more representative of the target image, and thus capturing more objects. However, the trend remains the same. Increasing the temperature decreases the memorization, but the default value of 100 is close to optimal as the trade-off between memorization and model utility is the best. Increasing the weight decay improves the model utility (indicated by the zero-shot accuracy) and decreases memorization. Weight decay of 0.3 gives near optimal trade-offs. Further increasing $wd$ to 1.0 results in a drastic decrease in model utility, and thus we do not include the results. Increasing the masking ratio from 0 to 0.5 significantly reduces the memorization but at the cost of model utility. While the optimal value of $mr$ would depend on the application and how much tolerance on the model utility loss is acceptable, we find that $mr = 0.3$ achieves a significant reduction in memorization while only moderately impacting the zero-shot accuracy, as shown in \autoref{fig:defense_impact_laion}. Any further increase in $mr$ beyond 0.5 would greatly sacrifice the model utility and thus is not recommended.

\section{Additional Results with Shutterstock}\label{sec:additional_results_ss}

\begin{figure*}[tb]
    \centering
    \begin{subfigure}[tb]{\textwidth}
    \centering
        \includegraphics[width=\textwidth]{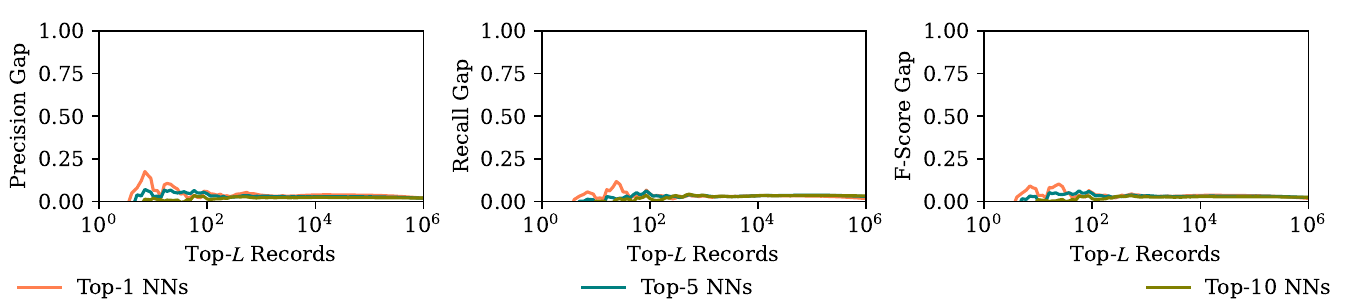}
        \caption{Records are sorted w.r.t. the minimum embedding distance between target caption and public images.}
        \label{fig:lab_inf_top_k_recs_ss_min_dist}    
    \end{subfigure}
    \begin{subfigure}[tb]{\textwidth}
    \centering
        \includegraphics[width=\textwidth]{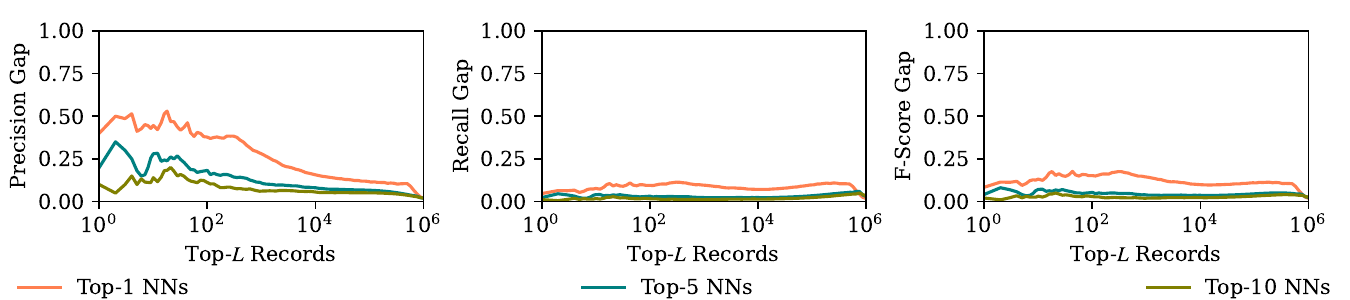}
        \caption{Records are sorted w.r.t. the decreasing number of correct object predictions for target model.}
        \label{fig:lab_inf_top_k_recs_ss_crct_pred}
    \end{subfigure}
    \caption{Sample-level memorization gap between target and reference models when predicting top-10 objects for different top-$L$ records. Models are trained on disjoint 10M subsets of Shutterstock data set for 200 epochs and SS-20M public set is used for the KNN test.}
    \label{fig:lab_inf_top_k_recs_ss}
\end{figure*}

Similar to the sample-level evaluation for models trained on filtered LAION data set in \autoref{sec:imagenet_sample}, we evaluate the gap in precision, recall and F-scores of top-$k$ records sorted with respect to the minimum embedding distance for models trained on Shutterstock data set when considering SS-20M as the public set for the nearest neighbor test. \autoref{fig:lab_inf_top_k_recs_ss_min_dist} shows the memorization gap of the top-$k$ records. We note smaller precision and recall gaps for this data set. This is due to two reasons: (a) nature of the data set--- Shutterstock data set has many similar images even after we do the caption-level deduplication (see \autoref{sec:setup}) so even the referece model performs well on this data set, and (b) the model hyper-parameter settings for this training set size is possibly sub-optimal--- the model zero-shot accuracy on ImageNet seems to be the highest at 20 epochs (18.16\%) and it slightly decreases till 200 epochs (17.49\%) when trained on 10M subset of Shutterstock data. \autoref{fig:lab_inf_top_k_recs_ss_crct_pred} shows the memorization gap when we sort the records in the decreasing order of the number of correct predictions made by the target model using the nearest neighbor test. As expected this gap is much higher than the previous case. Overall, the trends are similar to the filtered LAION experiments.

\begin{figure*}[tb]
    \centering
    \begin{subfigure}{\textwidth}
        \includegraphics[width=\linewidth]{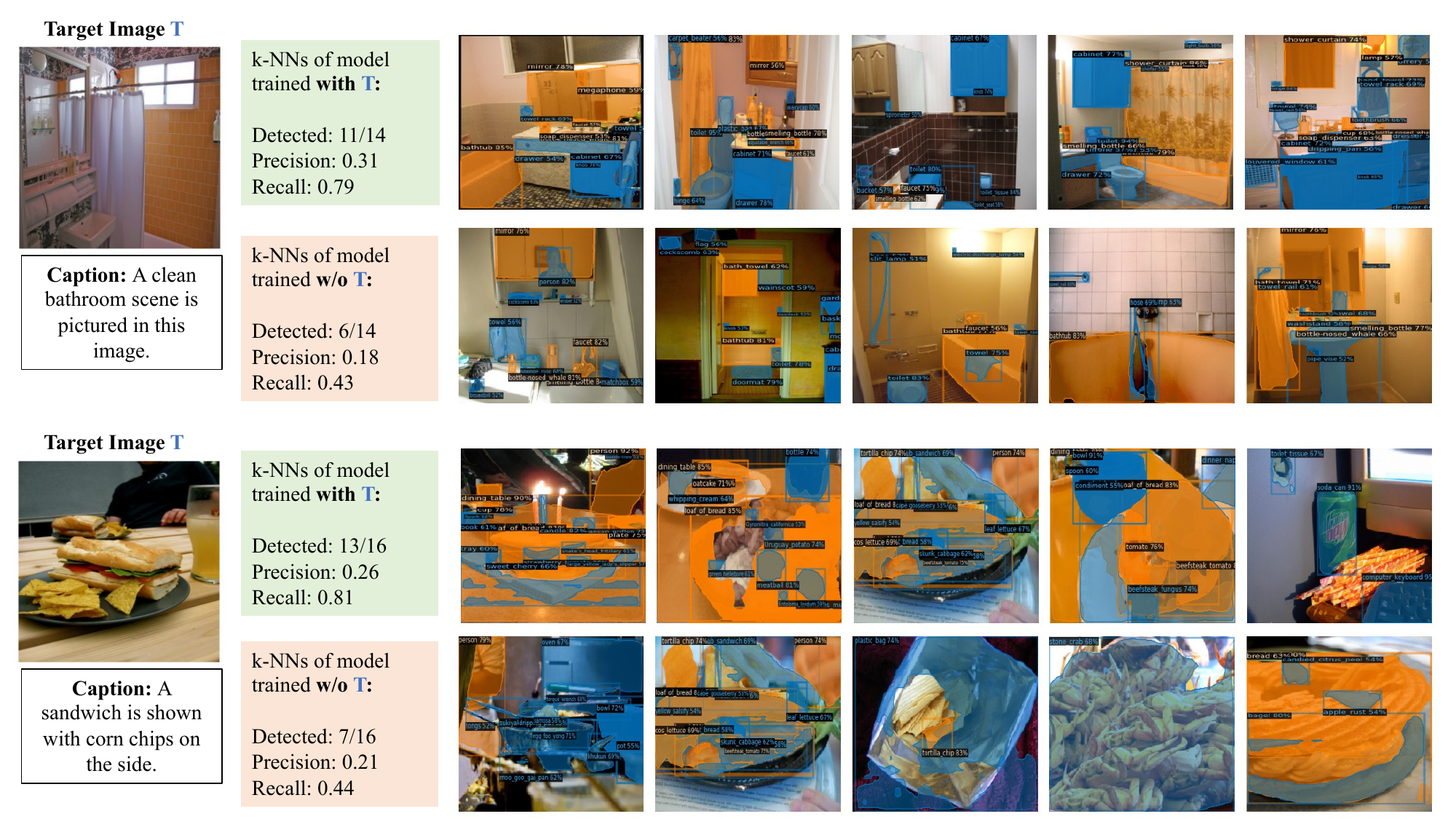}
    \end{subfigure}
    \begin{subfigure}{\textwidth}
        \includegraphics[width=\linewidth]{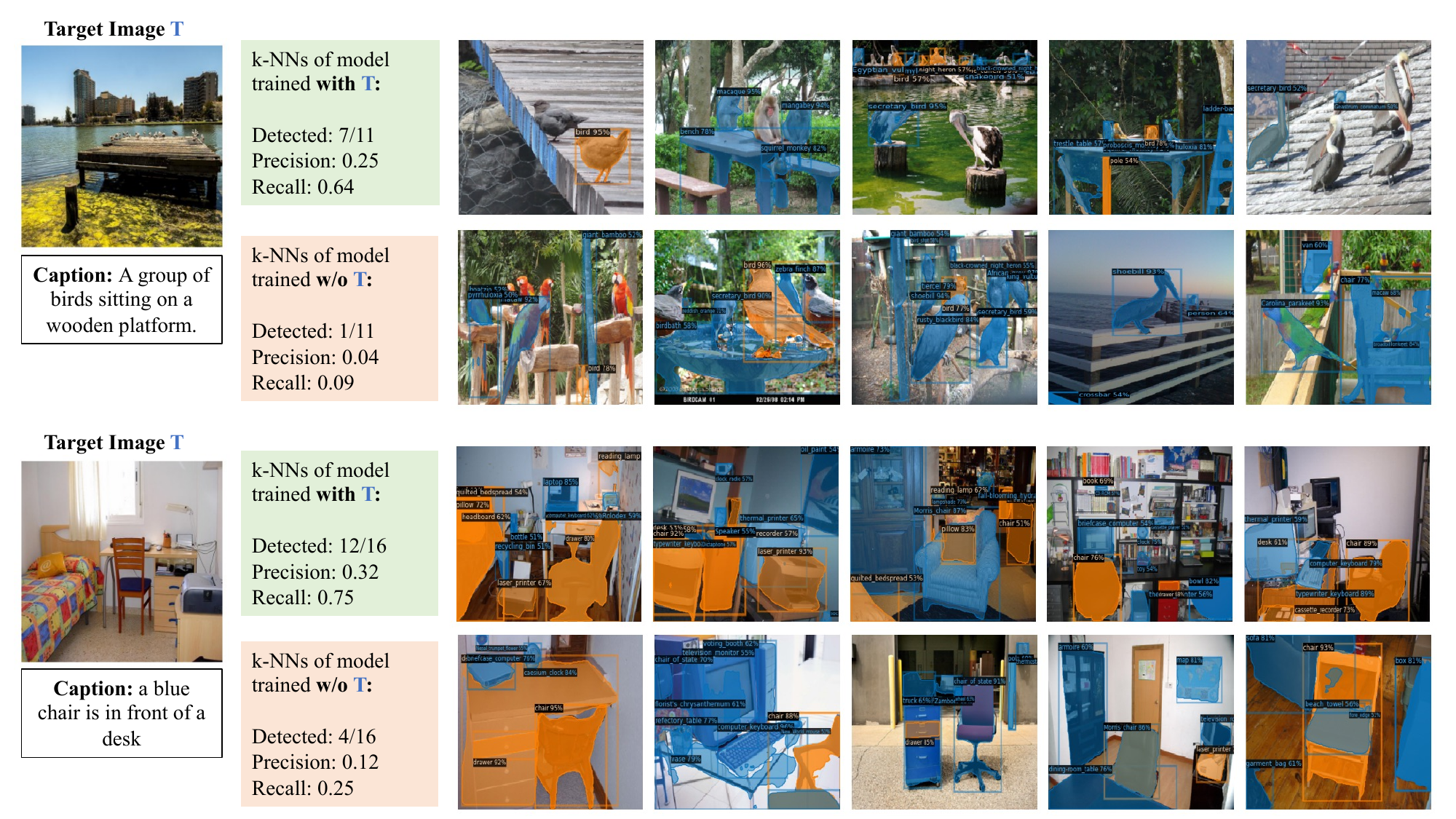}
    \end{subfigure}
    \caption{Additional examples showing \dejavu memorization. Target images are from COCO training set and the public images are from ImageNet data set. The objects annotated in {\color{orange} orange} are true positives, i.e., the ones present in the target image, and the objects annotated in {\color{blue} blue} are false positives.}
    \label{fig:additional_examples_coco}
\end{figure*}

\begin{figure*}[tb]
    \centering
    \begin{subfigure}{\textwidth}
        \includegraphics[width=\linewidth]{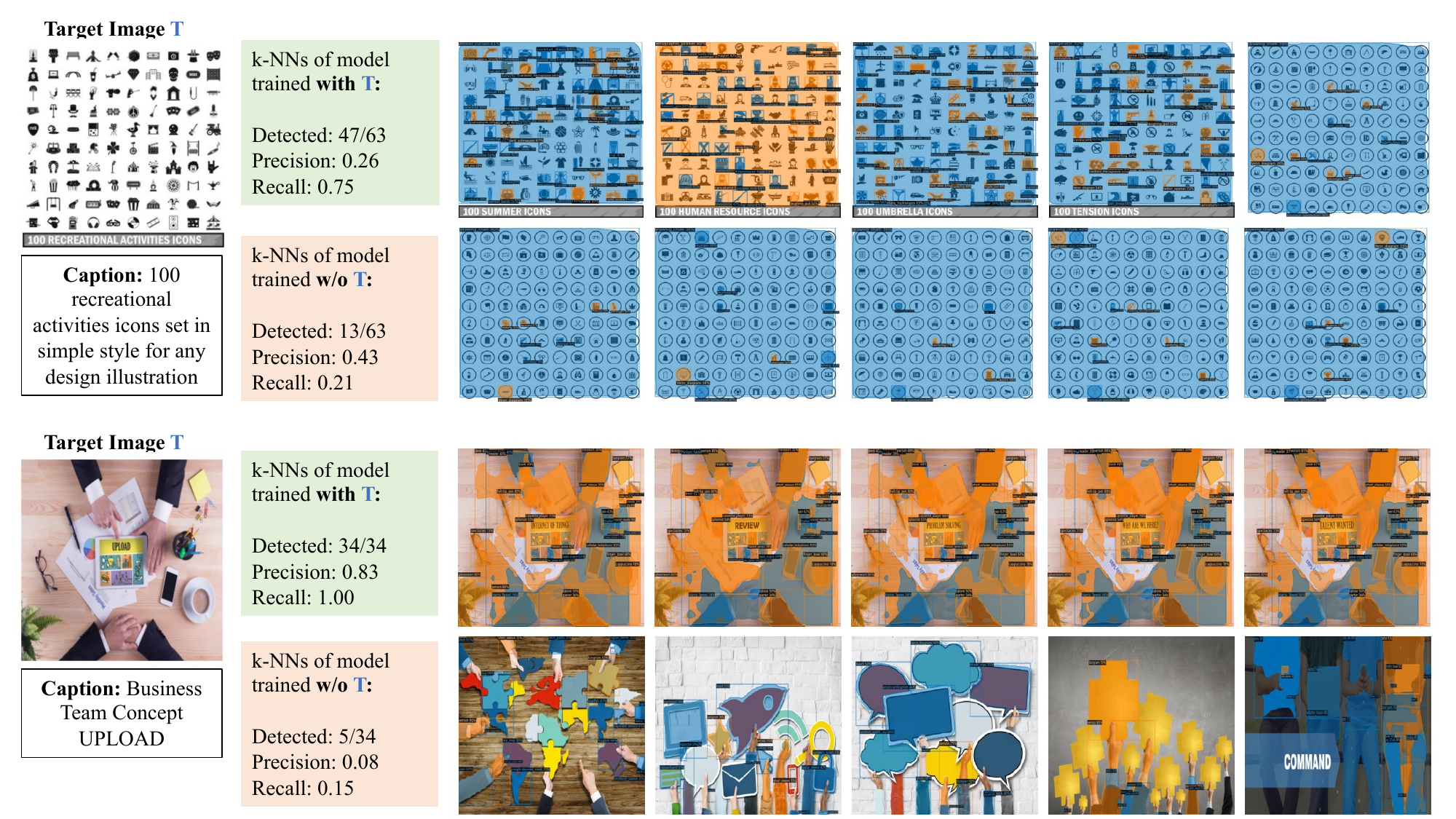}
    \end{subfigure}
    \begin{subfigure}{\textwidth}
        \includegraphics[width=\linewidth]{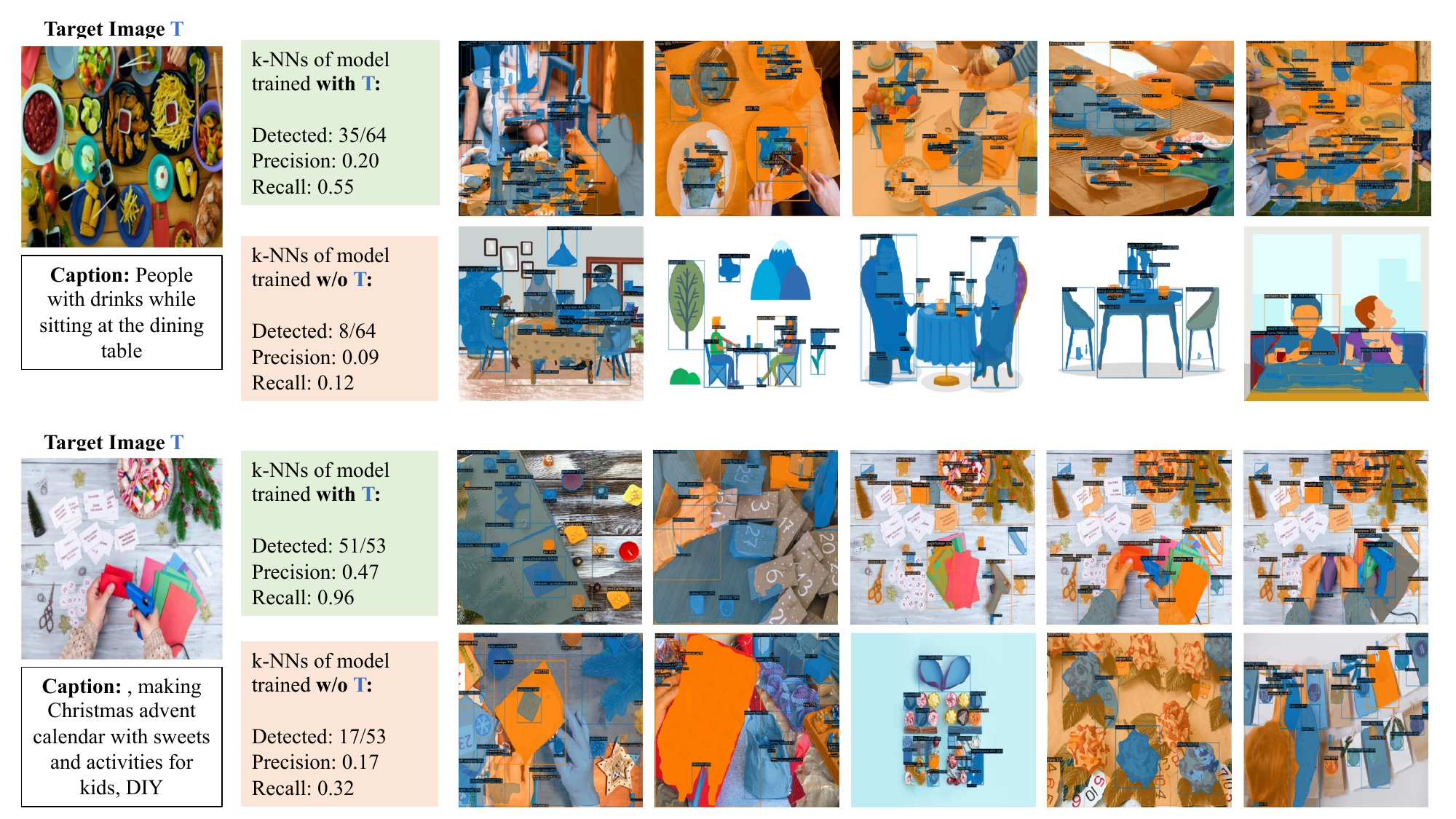}
    \end{subfigure}
    \caption{Additional examples showing \dejavu memorization. Target images are from Shutterstock training set and the public images are from SS-20M public set. The objects annotated in {\color{orange} orange} are true positives, i.e., the ones present in the target image, and the objects annotated in {\color{blue} blue} are false positives.}
    \label{fig:additional_examples_ss}
\end{figure*}

\end{document}